\documentclass[runningheads]{llncs}
\usepackage[T1]{fontenc}
\usepackage{graphicx}
\usepackage{rotating}
\usepackage{amssymb}
\usepackage{pdflscape}
\usepackage{makecell}

\begin{document}
	\title{Initial Steps in Integrating Large Reasoning and Action Models for Service Composition}
	\author{Ilche Georgievski\orcidID{0000-0001-6745-0063} \and
		Marco Aiello\orcidID{0000-0002-0764-2124}}
		\institute{University of Stuttgart, Stuttgart, Germany\\
			\email{\{ilche.georgievski,marco.aiello\}@iaas.uni-stuttgart.de} 
		}
		\authorrunning{Georgievski and Aiello}
		
		\maketitle              % typeset the header of the contribution
		\begin{abstract}
			Service composition remains a central challenge in building adaptive and intelligent software systems, often constrained by limited reasoning capabilities or brittle execution mechanisms. This paper explores the integration of two emerging paradigms enabled by large language models: Large Reasoning Models (LRMs) and Large Action Models (LAMs). We argue that LRMs address the challenges of semantic reasoning and ecosystem complexity while LAMs excel in dynamic action execution and system interoperability. However, each paradigm has complementary limitations--LRMs lack grounded action capabilities, and LAMs often struggle with deep reasoning. We propose an integrated LRM-LAM architectural framework as a promising direction for advancing automated service composition. Such a system can reason about service requirements and constraints while dynamically executing workflows, thus bridging the gap between intention and execution. This integration has the potential to transform service composition into a fully automated, user-friendly process driven by high-level natural language intent.
			
		\end{abstract}
		\section{Introduction}
		
		Automated service composition is the capability of a system to autonomously interpret a user request, identify suitable services available on a network, and then select, orchestrate, and execute services in an automated and unsupervised manner. Achieving automated service composition represents a central problem in service-oriented computing~\cite{papazoglou2003:soc}. As this is a complex problem, one can conceptualise it as a continuum of capabilities, ranging from Level 0 (no automation, entirely manual process) to Level 5 (full automation, a system autonomously handles the entire service composition lifecycle), in a similar fashion to autonomous driving~\cite{aiello2022:service-composition-challenge}. These levels provide a systematic approach to understanding and assessing the maturity of automated service composition and identifying specific challenges that must be addressed to achieve higher levels of automation. Currently, most established automated service composition techniques, which primarily rely on rule-based systems, semantic Web technologies, and AI planning systems, typically achieve around Level 2 automation. However, several critical challenges remain a significant barrier to moving automated service composition toward full automation (Level 5):
		
		\begin{itemize}
			\item \textbf{Challenge 1: Context understanding}. Existing approaches struggle to interpret service requirements and to account for implicit constraints that are not fully or formally specified in the user request.    
			\item \textbf{Challenge 2: Service integration}. The cognitive load of integrating heterogeneous services with different interfaces, data models, and behavioural patterns remains largely unaddressed. This integration often requires domain knowledge and technical expertise that existing systems have difficulty replicating.
			\item \textbf{Challenge 3: Service execution and adaptation}. When service compositions are deployed in dynamic environments, they frequently fail due to changing conditions or unexpected errors. The ability to monitor service execution, detect issues, and implement recovery strategies in real time remains largely unimplemented.
			\item \textbf{Challenge 4: Reasoning at scale}. As service ecosystems grow in size and complexity, reasoning about compatibility, substitutability, and optimal composition becomes significantly more difficult.
		\end{itemize}
		
		The rapid advancement of Large Language Models (LLMs) is dramatically transforming the research landscape across multiple domains, offering new approaches to problems that require human-level intelligence and flexibility. Within service-oriented computing, we have undertaken an early exploration of the potential of LLMs for automated service composition~\cite{aiello2023:llm-service-comopsition} and further discussed the paradigm shift that LLMs have set in motion for service composition~\cite{aiello2025:paradigm-shift}. Much of the research in this context primarily focuses on user request specification via sophisticated prompting techniques and service discovery, thanks to the vast knowledge base of LLMs. Current research also partially focuses on generating structured and possibly executable service compositions. These efforts effectively contribute to Challenges~1 and~2, while showing significant limitations in addressing Challenge~4. To the best of our knowledge, Challenge~3 still remains largely unexplored, as previously pointed out in~\cite{aiello2023:llm-service-comopsition} and illustrated later in Section~\ref{sec:soa}. These gaps stem from fundamental limitations in current LLMs: their general-purpose capabilities, which prioritise breath over depth in specific domains, shallow reasoning capabilities even when augmented with advanced prompting techniques and fine-tuned, and limited direct interaction capabilities with digital and physical environments. 
		
		Two paradigms that emerge from the recent developments in LLMs, namely Large Reasoning Models (LRMs) and Large Action Models (LAMs), present a promising tool to address all four challenges. LRMs are exemplified by models designed primarily for complex reasoning, abstract understanding, context-sensitive decision-making, and nuanced semantic tasks~\cite{besta2025:rlm}. LAMs are specialised models explicitly built or fine-tuned to coordinate and execute actions through external tools, APIs, or function-calling capabilities~\cite{wang2024:lam}. LRMs offer a solution to Challenges~1 and~4 by leveraging their exceptional capabilities in understanding natural language service descriptions, reasoning about service compatibility beyond simple interface matching, inferring implicit requirements and constraints, and handling the complexity of large service ecosystems through their ability to process vast contextual information. On the other hand, LAMs offer a solution to Challenges~2 and~3 through their ability to execute and adapt actions in changing environments, skill in interfacing with diverse systems through appropriate API calls, capacity to learn from execution feedback and improve over time, and error handling and recovery mechanisms.
		
		LRMs excel in high-level reasoning but typically lack reliable direct action execution capabilities. Conversely, LAMs can reliably execute tasks and interact with external systems but often exhibit shallow reasoning or limited adaptability to novel contexts. The combination of these two paradigms represents a new approach that could bridge critical gaps in automated service composition research.
		
		We envision the LRM-LAM integration creating systems that can compose complex service workflows by reasoning deeply about requirements, constraints, and service characteristics while dynamically executing the necessary actions with reliability and adaptability. This vision will transform service composition from a semi-automated process to one where developers and end-users can express high-level intentions in natural language, leaving the reasoning about request interpretation, service selection, composition, and execution to AI systems capable of understanding both the "why" and the "how" of service composition.
		
		The rest of the paper is organised as follows. We provide background information in Section~\ref{sec:fundamentals}. Then, we overview the state of the art in Section~\ref{sec:soa}. The proposed architectural framework is the object of Section~\ref{sec:architecture}. Final considerations are contained in Section~\ref{sec:conclusions}.

		\section{Background}
		\label{sec:fundamentals}
		
		To provide the background for the present proposal, we first recall how service composition works in terms of stages in a pipeline. Then, we recall the main features of LLMs, and their specialisation as LRMs and LAMs.

		\subsection{Automated Service Composition Pipeline}
		
		A service composition combines services together to achieve a given objective. Automated service composition is the computational process of finding, selecting, and organising services such that the resulting service composition achieves the given objective. Considering the phases automated service composition goes through, we can distinguish six phases that can be organised in a pipeline-like fashion. For simplicity, we consider the phases to be executed in a sequential order, whereas in some cases, some phases may happen in parallel. 
		
\begin{figure}[t]
			\centering
			\includegraphics[width=1\linewidth]{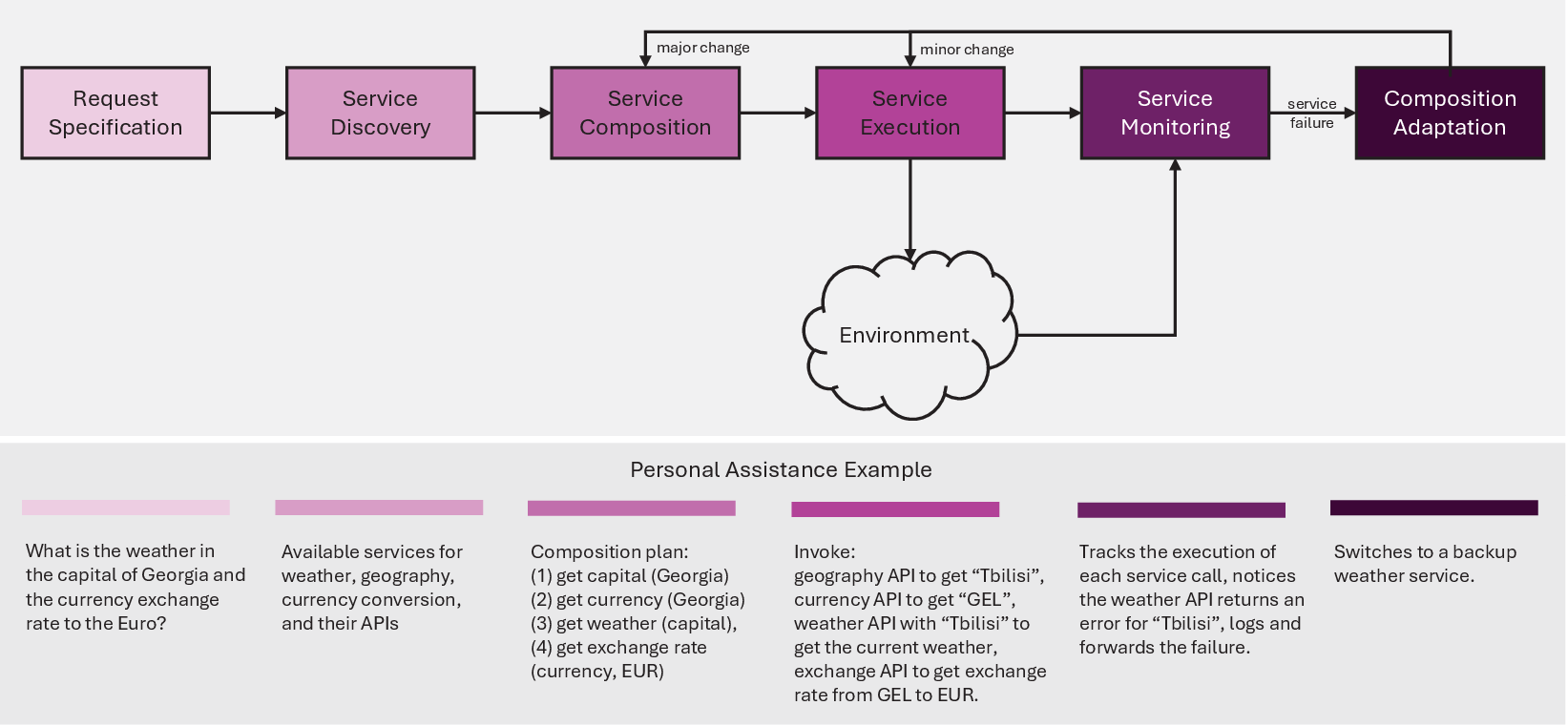}
			\caption{Automated service composition pipeline with illustrative example from the personal assistance domain.}
			\label{fig:asc-pipeline}
		\end{figure}

		Fig.~\ref{fig:asc-pipeline} shows the automated service composition pipeline with six phases: request specification, service discovery, service composition, service execution, service monitoring, and composition adaptation. In the first phase, the user request is specified. This can be a direct input from the user, or the system can observe and specify the user's objectives. In the second phase, the system searches and identifies available services that could potentially fulfill parts of the request. Traditionally, this phase involves querying service registries, repositories, and marketplaces to find available services. In the third phase, after a pool of relevant services is obtained, appropriate services are selected and composed such that they adhere to the request specification. This requires resolving service dependencies, ensuring data compatibility, and considering quality attributes, while verifying that the resulting composition meets all the requirements from the request specification. In the fourth phase, the resulting service composition is deployed on the corresponding infrastructure, and service instances are executed in the specified order in the environment, be it digital or physical. In the fifth phase, the system monitors the behaviour of individual services and the service composition as a whole during runtime and gathers relevant observations. In the final phase, the system may adapt the service composition in response to observations from the monitoring phase. Service failures, environmental changes, context changes, changes in requirements, etc, may trigger adaptations. The strategies to adapt include adjusting the existing service composition, for example, by substituting a service or requesting a recomposition. 
		
		Fig.~\ref{fig:asc-pipeline} also includes a phase-wise example from the personal assistance domain, demonstrating a possible realisation of each step of the pipeline.
		
		\subsection{Large Language Models (LLMs)}
		Large Language Models (LLMs) are general-purpose models that are trained on large-scale amounts of text with the task of predicting meaningful sequences of tokens~\cite{zhao2023:llm-survey}. Additionally, LLMs have \textit{emergent abilities}~\cite{wei2022:emergent-abilities}, such as in-context learning~\cite{dong2022:in-context-survey} and role-playing~\cite{shanahan2023:role-play}. These abilities enable these general-purpose LLMs to excel in understanding human language and generating coherent responses, such as text, code, and other similar outputs. 
		
		The question of whether LLMs have the potential to solve complex real-world problems through abstract and logical reasoning has been of particular interest and debate. A notable concept in this context is the concept of \textit{thought}, which is a sequence of tokens that represents the intermediate steps in human-like reasoning processes. The initial approach based on this concept is the Chain of Thought prompting technique~\cite{wei2022:cot}, which has led to the development of more advanced prompting techniques, such as Tree of Thought~\cite{yao2023:tot}. By incorporating such intermediate steps, it was expected for LLMs to move to exhibit reasoning processes while keeping their auto-regressive token generation. Despite this and fine-tuning, the reasoning capabilities of LLMs remain shallow, as they rely on the next-token prediction and prompting techniques rather than explicit reasoning and problem-solving~\cite{valmeekam2022:llm-cant-plan,besta2025:rlm,xu2025:lrm}. Another research direction investigates LLM-modulo frameworks, that is, the combination of LLMs with external verifying, solving, and other symbolic components for improved reasoning and decision-making, showing more promise than thought-based prompting techniques and fine-tuning~\cite{yao2023:react,kambhampati2024:llm-modulo}. 
		
		Similarly, the capabilities of LLMs to directly interact with or manipulate the digital or physical world are limited. One line of research that looked into this is related to the concept of \textit{language agents}, which are agents that use LLMs to interact with external tools or environments with the objective of perceiving the environment, reasoning, acting, and receiving feedback~\cite{zeng2023:language-agents,chen2023:fireact}. Most LLMs show poor performance when used as agents. Fine-tuning of these models may be a possible solution to improve the performance. A few works have fine-tuned LLMs for Web navigation, e.g.,~\cite{yao2022:webshop}, API tool use, e.g.,~\cite{schick2023:toolformer}, or Google Search~\cite{chen2023:fireact}.

		\begin{figure}[t]
			\centering
			\includegraphics[width=1\linewidth]{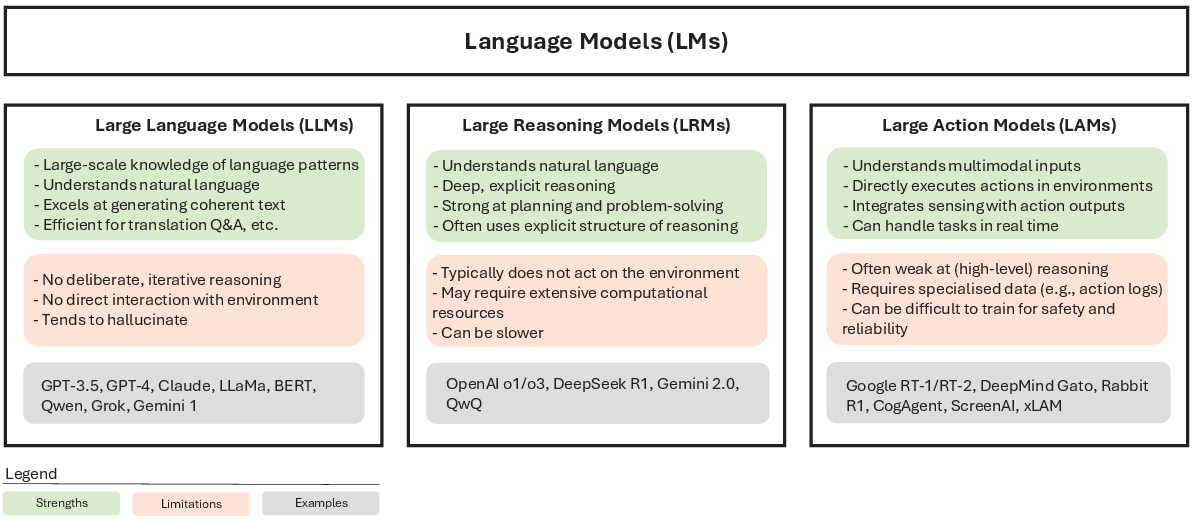}
			\caption{Language Model Types}
			\label{fig:llm-lrm-lam}
		\end{figure}

		\subsection{Large Reasoning Models (LRMs)}
		
		Large Reasoning Models (LRMs) combine the capability of knowledge-rich LLMs with the capabilities of reasoning techniques, such as reinforcement learning~\cite{xu2025:lrm,besta2025:rlm}. These capabilities allow for the exploration of large solution spaces, evaluation of multiple strategies, and iterative optimisation of solutions. The result is a new class of models that can engage in explicit and deliberate reasoning, providing improved context understanding, intuitive pattern recognition, nuanced problem-solving, and powerful decision-making. 
		
		The foundational architecture of LRMs consists of three main pipelines: inference, training, and data generation~\cite{besta2025:rlm}. The inference pipeline directly handles user requests in real-time. The pipeline applies deployed models that result from the training pipeline.  The training pipeline takes collected data and uses it to train new models, update existing ones, and improve the overall performance of the LRMs. It may use techniques such as reinforcement learning and AI planning. The data generation pipeline has a similar architecture to the inference pipeline but with a key difference in objective: it operates independently of user requests. It may run continuously or on a scheduled basis to systematically create new training examples across multiple domains, which are then fed into the training pipeline. By generating data across multiple domains, LRMs can develop more robust and generalisable capabilities. Together, these pipelines create a self-improving system.
		
		The key limitations of LRMs are that they are not trained to perform actions in environments and may incur high and unpredictable costs while being slow in completing the tasks, as shown for the OpenAI's o1 model in~\cite{valmeekam2025:lrm-o1}.

		\subsection{Large Action Models (LAMs)}
		
		Large Action Models (LAMs) are an emerging class of AI models that do not only generate text but also generate and execute actions in (interactive) environments. LAMs are designed to perceive situations and carry out a complex course of action in response to instructions. LAMs also combine the capabilities of knowledge-rich LLMs but they do so with the capabilities of other predictive models and real-time analytics for interaction with or manipulation of the physical or digital environment. LAMs typically accept multi-modal inputs, such as natural language, visual, or sensor data, and produce sequences of actions, such as robot motions, game plays, and user interface clicks. To achieve this, LAMs integrate user intention interpretation, action generation grounded in the environment, action execution in the environment, and reacting to environmental changes~\cite{wang2024:lam}. LAMs often rely on learned patterns of behaviour and may have limited ability to reason and plan in complex situations. 
		
		LAMs assume the existence of an agent framework, which serves as an operational platform. Such a framework may include agents for gathering observations, utilising tools, and enabling feedback loops. Once a LAM has been trained, the model must be incorporated into the agent framework so that it can effectively execute actions and adapt the behaviour based on real-time feedback.
		
		Based on the LAM development process described in~\cite{wang2024:lam}, to illustrate their working and use, we extract a foundational architecture of LAMs that consists of five pipelines: training, evaluation, integration, inference, and data generation. The training pipeline uses techniques such as supervised fine-tuning to develop a model that can perform actions effectively. The evaluation pipeline focuses on offline evaluation, which assesses the model's performance using static datasets in a controlled environment, and an online evaluation, which assesses the model in real-world conditions. The integration pipeline connects the trained model with the agent framework. This integration transforms the language model into an actionable system. The inference pipeline uses the trained and integrated model to process requests, execute appropriate actions via the agent framework, interact with external tools and environments, and deliver results. The data generation pipeline has a similar purpose to LRM's data generation pipeline and creates training examples in relation to requests, environmental context, actions, and other relevant information.
		
		LAMs are being developed across various domains, mainly in robotics and software-based interactive systems. Robotics is a natural application for LAMs as they process sensor inputs and output control actions for robots, allowing them to perform tasks in the physical world, e.g.,~\cite{brohan2023:rt}. Beyond robotics, LAMs are being applied to enable embodied intelligence in virtual environments, such as video games, simulators, or virtual worlds, where agents can take actions, such as controlling a game character, navigating a simulated household, or coordinating with humans in a virtual world (for example, see~\cite{wang2023:voyager}). Another growing area for LAMs is in digital assistants and agents that operate software or the Web on behalf of users, e.g.,~\cite{niu2024:screenagent}. Instead of a physical body, these agents act through software interfaces by clicking buttons, typing, opening apps, or calling APIs.
		
		Fig.~\ref{fig:llm-lrm-lam} summarises the strengths and limitations of LLMs, LRMs, and LAMs, and provides examples of models in each category.

		\section{State of the Art}
		\label{sec:soa}
		
		There are a handful of studies that investigate the use of LLM for automated service composition. Following our presentation of fundamentals, we analyse these existing works from the perspective of their focus on automated service composition phases and the kind of LLMs they use, as well as which tools, methods, and techniques they use in addition to LLMs. Table~\ref{tab:llm_phases} gives a summary of the former, and Table~\ref{tab:llm_properties} gives a summary of the latter.
		
		Table~\ref{tab:llm_phases} shows that almost all studies assume the request to be specified in natural language. However, more than half of the studies expect the user to provide technical or expert content, such as OpenAPI specifications of services. A few studies expect the users to follow a template when specifying the service composition request. Almost all studies focus on service discovery either by using the knowledge base of LLMs or by providing databases or Retrieval Augmented Generation (RAG) bases with service specifications, such as APIs and OpenAPI documentation. Almost half of the studies also deal with the service composition phase, tasking the respective LLM to select and order discovered services such that the user request is satisfied. Many of the approaches employ iterative prompting to refine the service composition result, as it is often incorrect or invalid. A few approaches also execute the resulting service composition for the purpose of testing the approach and not the actual deployment of the service composition. Only one approach focused on all phases of automated service composition, where the composed services are executed, and their execution is monitored in case adjustments of the computed service composition are needed. This approach interleaves the service composition phase with the service execution and monitoring phases.
		
		\begin{table}[t]
			\centering
			\caption{Use of LLMs across automated service composition phases. $\bigodot$ indicates an existing work contributes to the corresponding phase, whereas \texttimes indicates no contribution. NL stands for natural language, and TC for technical content.}
			\label{tab:llm_phases}
			\footnotesize
			\begin{tabular}{|l|l|c|c|c|c|c|c|}
			\hline
			\textbf{Work} & \textbf{\makecell{Request \\ Specification}} & \textbf{\makecell{Service \\ Discovery}} & \textbf{\makecell{Service \\ Composition}} & \textbf{\makecell{Service \\ Execution}} & \textbf{\makecell{Service \\ Monitoring}} & \textbf{\makecell{Service \\ Adaptation}} \\
			\hline
			\cite{wang2021:servicebert} & \texttimes & $\bigodot$ & \texttimes & \texttimes & \texttimes & \texttimes \\
			\hline
			\cite{aiello2023:llm-service-comopsition} & \makecell[l]{Prompt \\ NL} & $\bigodot$ & $\bigodot$ & $\bigodot$ & \texttimes & \texttimes \\ \hline
			\cite{huang2023:llm-service-discovery} & \makecell[l]{Template prompt \\ NL + TC} & $\bigodot$ & \texttimes & \texttimes & \texttimes & \texttimes \\
			\hline
			\cite{pesl2023:llm-service-composition} & \makecell[l]{Prompt \\ NL + TC} & $\bigodot$ & $\bigodot$ & \texttimes & \texttimes & \texttimes \\
			\hline
			\cite{song2023:restgpt} & \makecell[l]{Prompt \\ NL} & $\bigodot$ & $\bigodot$ & $\bigodot$ & $\bigodot$ & $\bigodot$ \\
			\hline
			\cite{bianchini2024:llm-service-discovery} & \makecell[l]{Prompt \\ NL + TC} & $\bigodot$ & \texttimes & \texttimes & \texttimes & \texttimes \\
			\hline
			\cite{kotstein2024:llm-service-discovery} & \makecell[l]{Prompt \\ NL + TC} & $\bigodot$ & \texttimes & \texttimes & \texttimes & \texttimes \\
			\hline 
			\cite{pesl2024:composito} & \makecell[l]{Template prompt \\ NL + TC} & \texttimes & $\bigodot$ & \texttimes & \texttimes & \texttimes \\
			\hline
			\cite{zhang2024:llm-agent-service-composition} & \makecell[l]{Template prompt \\ NL} & $\bigodot$ & $\bigodot$ & $\bigodot$ & \texttimes & \texttimes \\
			\hline
		\end{tabular}
	\end{table}
	
	Table~\ref{tab:llm_properties} shows that almost all studies use general-purpose LLMs, and only a few studies use domain-specific LLMs, that is, LLMs aimed for coding, and one study also uses a language understanding model (without generation capabilities). The models used range from pre-trained models, such as GPT-3.5, GPT-4, Llama, Mixtral, and Qwen, to fine-tuned models, such as CodeBERT and RoBERTa, to custom models such as ServiceBERT. In all but one study, the approaches use tools, methods, and techniques in addition to the LLMs. In half of the studies, OpenAPI is used to guide LLMs into discovering and creating appropriate service descriptions, RAG is used to expand the knowledge base of LLMs typically with information about service descriptions, iterative prompting is used to improve the LLM result for service composition, and question answering is used to refine the initial user prompt into a query with suitable format and content.
	
	The table also includes two qualitative dimension, namely feasibility and impact. Feasibility reflects the practical effort required to implement an approach, considering factors such as model availability, training requirements, architectural complexity, and integration effort. Approaches that rely on publicly accessible, off-the-shelf LLMs, and standard APIs are considered highly feasible, while those requiring fine-tuning, retrieval infrastructure, or component orchestration are assigned lower feasibility. Impact refers to the expected contribution of each approach toward achieving automated service composition, including the extent to which the approach enables request interpretation, service discovery, composition, execution, or adaptation. Higher impact is attributed to approaches that support multiple phases of the automated service composition pipeline or demonstrate advanced automation capabilities for a single phase. The assigned labels are approximate and based on our interpretation of each work's scope and reported contributions.
	
	\begin{table}[t]
		\centering
		\caption{LLM types, models, and additional tools/methods used in existing works.}
		\label{tab:llm_properties}
		\footnotesize
		\begin{tabular}{|l|l|l|l|l|l|}
			\hline
			\textbf{Work} & \textbf{LLM Type} & \textbf{Model(s)} & \textbf{Tools/Methods} & \textbf{Feasibility} & \textbf{Impact}\\
			\hline
			\cite{wang2021:servicebert} & Domain-specific & ServiceBERT & - & Medium & Low \\
			\hline
			\cite{aiello2023:llm-service-comopsition} & General-purpose & GPT-3.5 & Iterative prompting & High & Low \\ 
			\hline
			\cite{huang2023:llm-service-discovery} & General-purpose & GPT-3.5 & \makecell[l]{Knowledge Graphs \\ Question Answering \\ Iterative prompting} & Medium & Medium \\
			\hline
			\cite{pesl2023:llm-service-composition} & General-purpose & GPT-4 & \makecell[l]{Iterative prompting \\ OpenAPI \\ Python code} & Medium & High \\
			\hline
			\cite{song2023:restgpt} & General-purpose & text-davinci-003 & \makecell[l]{RAG \\ OpenAPI \\ Python code} & Low & High \\
			\hline 
			\cite{bianchini2024:llm-service-discovery} & General-purpose & Mistral 7B instruct & \makecell[l]{RAG \\ OpenAPI} & Low & Medium\\
			\hline
			\cite{kotstein2024:llm-service-discovery} & \makecell[l]{Domain-specific \\ Language under.} & \makecell[l]{CodeBERT, \\ RoBERTa \\ both fine-tuned} & \makecell[l]{Question Answering \\ OpenAPI} & Low & Medium \\
			\hline
			\cite{pesl2024:composito} & \makecell[l]{General-purpose \\ Domain-specific} & \makecell[l]{Llama 2, \\ CodeLlama, \\ Llama 3, Mixtral, \\ Qwen, GPT-3.5, \\ GPT-4} & \makecell[l]{OpenAPI \\ JSON} & Medium & Low \\
			\hline
			\cite{zhang2024:llm-agent-service-composition} & General-purpose & GPT-4, GPT-3.5 & \makecell[l]{RAG \\ DFS decision \\ tree algorithm} & Medium & High \\ 
			\hline
		\end{tabular}
	\end{table}	
	
	\section{Architectural Framework for End-to-End Automated Service Composition}
	\label{sec:architecture}
	
	We observe that LRMs and LAMs show distinct yet complementary strengths and limitations. Importantly, this complementarity aligns well with the requirements for fully automating the service composition pipeline. The idea is to envision LRMs functioning as the "brain" of automated service composition, understanding the user request for composition, determining which services to compose and how to compose them to satisfy the request in the best way, and LAMs serving as the "body", executing the computed service composition in real-world environments. Thus, integrating LRMs and LAMs forms a synergetic system capable of addressing the complete lifecycle of automated service composition. 
	
	To operationalise this synergy, we propose an architectural framework that distinguishes between three essential components: an inference phase, a coordination layer, and a training phase for automated service composition. The inference phase is central to the proposed approach, as it encompasses the tasks directly responsible for achieving automated service composition. This phase consists of three interconnected layers, each corresponding to distinct segments of the automated service composition pipeline. Although these layers are conceptually presented in a sequential manner, practical implementations would involve dynamic interactions, with AI systems interweaving LRM and LAM capabilities to ensure intelligent, flexible, and adaptable service composition and execution. 
	
	The training phase is essential for continuous improvement, allowing language models to be trained on domain-specific datasets and fine-tuned with insights generated from previous service computation and execution cycles. This creates a feedback loop that enhances automated service composition performance over time.
	
	The coordination layer serves two essential functions. First, it coordinates the operation of the inference layers, enabling information flow and continuous adaptation. For example, the middleware can maintain a consistent representation of the composition state between the reasoning (Layer~2) and execution (Layer~3) layers. It can also route execution results from Layer~3 back to Layer~2, enabling iterative and reflective refinement based on real-world performance. Second, the coordination layer can be responsible for systematically capturing, processing, and generating data from each inference layer for the training phase. For example, the \textit{Composition Data Generation} component is responsible for documenting and storing the logic and rationale for the selected composition, and the \textit{Execution Data Generation} component for documenting and storing service failure patterns and successful recovery strategies. 
	
\begin{sidewaysfigure}
		\centering
		\includegraphics[width=1\linewidth]{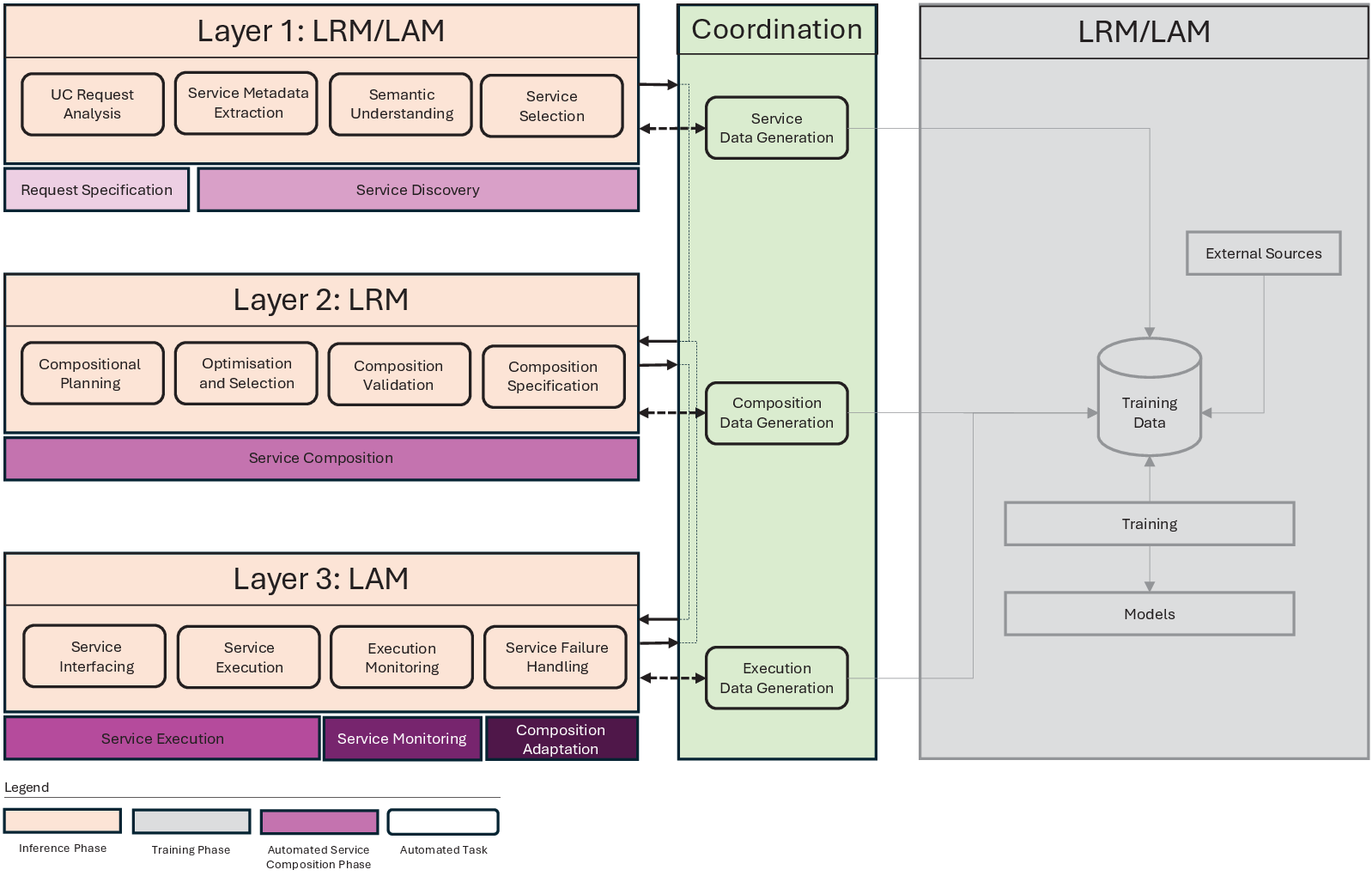}
		\caption{Architectural Framework}
		\label{fig:conceptual-model}
	\end{sidewaysfigure}

	Fig.~\ref{fig:conceptual-model} illustrates this architectural framework. Next, we present each layer involved in the inference phase.
	
	\subsection{Layer 1: Request Analysis \& Service Discovery}
	The \textit{Request Analysis \& Service Discovery Layer} leverages LRMs' and LAMs' capabilities to understand the user composition request and discover relevant services. This layer covers the first two automated service composition phases. The first phase is accomplished by the \textit{User Composition (UC) Request Analysis} task, where an LRM interprets the user requirements expressed in natural language and extracts functional needs, non-functional constraints, and other implicit constraints and expectations.
	
	The second phase is accomplished by performing three key tasks:
	\begin{itemize}
		\item \textit{Service Metadata Retrieval}, which uses a LAM to extract relevant service descriptions, including capabilities, API specifications, and metadata.
		\item \textit{Semantic Understanding}, which uses an LRM to create a semantic representation of service capabilities beyond API specifications, understanding the purpose, limitations, and compatibility of available services. This is necessary to establish semantic relationships between the UC request and service capabilities.
		\item \textit{Service Selection}, which uses an LRM to infer potential matches even with incomplete information, rank service candidates possibly using multi-criteria reasoning, and select the best service candidates. 
	\end{itemize}
	
	\subsection{Layer 2: Service Composition}
	Once services have been discovered through the combined LRM/LAM approach, an LRM can handle the service composition phase in terms of how to connect available and relevant services into an optimal and valid composition. This phase has an entire layer, called \textit{Service Composition Layer}, which consists of performing three key tasks:
	
	\begin{itemize}
		\item \textit{Composition Planning}, which uses an LRM to reason about service combinations, generating candidate compositions that satisfy the requirements and constraints analysed in the UC request.
		\item \textit{Composition Optimisation and Selection}, which computes quality metrics per candidate composition, identifies potential performance bottlenecks, and selects the best composition candidate.
		\item \textit{Composition Validation}, which checks the logical correctness of the selected composition and the satisfaction of all the constraints by the composition.
	\end{itemize}
	
	\subsection{Layer 3: Service Execution \& Adaptation}
	The \textit{Service Execution \& Adaptation Layer} employs LAMs to transform the computed service composition into a concrete specification with service instances, execute those instances, monitor their execution, and take appropriate steps when service execution fails. This layer covers two automated service composition phases, namely, service execution and service failure handling. The layer includes four key tasks:
	\begin{itemize}
		\item \textit{Composition Specification}, which uses a LAM to create a concrete workflow specification for the selected service composition.
		\item \textit{Service Interfacing}, which handles the mechanics of API calls, authentication, and data format conversions required to interact with heterogeneous services.
		\item \textit{Service Execution}, which invokes services in the correct order based on the workflow specification and determines optimal resource allocation for service execution.
		\item \textit{Service Execution Monitoring}, which tracks the execution state of each service, verifies outputs against expected outcomes, and detects deviations and failures in real-time.
		\item \textit{Service Failure Handling}, which determines root causes for failures, assesses failure types (e.g., transient or permanent), generates recovery strategies (e.g., retry, replacement services, reorder execution sequence, composition refinement), and restores the service composition to a consistent state after failures.
	\end{itemize}

	\section{Conclusion and Outlook}
	\label{sec:conclusions}
	
	We explored the possibility of achieving automated service composition by examining advanced language models beyond LLMs. In particular, we looked at the emerging paradigms of LRMs and LAMs as key components for building powerful solutions for fully automated service composition. We proposed an architectural framework that synergises LRMs and LAMs, leveraging their ability to understand and interpret natural language requests, their knowledge of service ecosystems to find available services, their reasoning capabilities for effective and optimal service combination, and their ability to execute services while detecting potential anomalies through feedback from the execution environments. 
	
	The envisioned general framework provides both a conceptual blueprint for developing potential solutions through its inference phase and an interface for integrating generated service composition and execution data into training for new models, as well as fine-tuning existing ones. We put an emphasis on the inference phase as a critical element that would facilitate the integration of our proposal into service composition solutions capable of achieving full automation. This framework is designed not merely as a linear pipeline but as a dynamic and cyclic architecture. The phases of discovery, composition, execution, and adaptation are interconnected through feedback loops and coordination mechanisms. This cyclicity allows the system to continuously refine its understanding of the environment, improve compositions over time, and learn from past successes and failures. The coordination layer plays a key role in enabling this interaction, supporting knowledge sharing and performance tracking between reasoning and action layers, and facilitating the continuous training of models with newly generated data.
	
	The presented architectural framework is aligned with the automated service composition pipeline to ensure full lifecycle coverage, aiming to enable intelligent and adaptive service composition across all pipeline phases.  In this way, the architecture not only serves as our core contribution but also offers a reflected research roadmap. We envision several lines of future work. First, evaluation and adaptation of existing LRMs and LAMs specifically for automated service composition are needed to identify their strengths, limitations, and necessary improvements for addressing the various tasks specified in the three layers of our architectural framework. Second, the development of prototypes of the conceptual architectural framework to validate and demonstrate its feasibility. Third, investigation of the effectiveness and reliability of prototypes integrating LRMs and LAMs across diverse service domains, focusing on context understanding, integrating heterogeneous service ecosystems, reasoning for valid and optimal service compositions at scale, and dynamically adapting service compositions in real-time as responses to identified service execution anomalies. Furthermore, exploration of how cycles of discovery, reasoning, execution, and adaptation can be orchestrated in flexible and context-aware ways. Next, enabling prototypes to support multi-modal input and design mechanisms for explaining composition decisions. Lastly, although current state-of-the-art works do not implement the proposed architecture, several can serve as starting points for realising the architecture's components. For example, approaches using general-purpose LLMs for service discovery or service composition can be extended or repurposed in the development of the corresponding architecture's components. 
	
	In summary, we believe that the integration of LRMs and LAMs marks an important step forward in service composition research. By combining reasoning and action capabilities in a unified architecture, we envision AI systems that can autonomously manage the full spectrum of composition tasks, transforming the current semi-automated landscape into a truly intelligent and adaptive service composition ecosystem.
	
	\bibliographystyle{splncs04}
	\bibliography{lit}

\end{document}